\documentclass[letterpaper, 10 pt, conference]{ieeeconf}  
\pdfoutput=1

\IEEEoverridecommandlockouts                              

\usepackage{graphics} 
\usepackage{subfigure}
\usepackage{epsfig} 
\usepackage{amsmath} 
\usepackage{amssymb}  
\usepackage[T1]{fontenc}
\usepackage{bm}
\usepackage{cite}
\usepackage{booktabs}
\usepackage{multirow}
\usepackage{diagbox}
\usepackage{threeparttable}
\usepackage{colortbl}
\usepackage{pifont} 
\usepackage{relsize}
\definecolor{linkblue}{rgb}{0, 0.19, 0.32}
\usepackage[colorlinks, linkcolor=black]{hyperref}
\hypersetup{
    colorlinks=true, 
    urlcolor=linkblue
    }

\usepackage{caption}
\newcommand \transpose {\mathsf{T}} 

\newcommand{\std}[1]{{\smaller$\pm$#1}}
 
\title{\LARGE \bf
CoorGrasp: Coordinated Contact Control for \\Adaptive Dexterous Grasping Under Uncertainty
}

\author{Anonymous Author(s)}

\author{Mingrui Yu, 
Yongpeng Jiang, 
Yongyi Jia,
Yi Ren,
and Xiang Li\textsuperscript{$\dagger$}
\thanks{
\textsuperscript{$\dagger$}Corresponding author: \texttt{xiangli@tsinghua.edu.cn}.}%
\thanks{This work was supported in part by the Brain Science and Brain-like Intelligence Technology-National Science and Technology Major Project under Grant 2021ZD0201404, in part by the National Natural Science Foundation of China under Grant 62461160307 and 623B2059, in part by the Fundamental and Interdisciplinary Disciplines Breakthrough Plan
of the Ministry of Education of China under Grant JYB2025XDXM208, and in part by the BNRist project under Grant BNR2024TD03003.}%
}%

\begin{document}

\maketitle
\pagestyle{empty}  
\thispagestyle{empty} 

\begin{abstract}
While recent research has focused heavily on dexterous grasp pose generation, less attention has been devoted to the execution of planned grasps. Under shape and position uncertainty, open-loop execution often yields uncoordinated contacts, causing undesired in-hand object motion and even grasp failures.
To address this, this paper proposes a tactile-driven model predictive controller for adaptive and delicate execution of diverse dexterous grasps.
Our approach emphasizes multi-contact coordination across both approaching and grasping phases, with three key novelties: (i) coordination-aware phase separation, (ii) arm–hand coordination to compensate for position errors, and (iii) adaptive force coordination to increase contact forces in a balanced manner.
An analytical model is employed to relate contact forces to robot joint motions for predictive control.
Our formulation imposes no restrictions on grasp types or contact configurations and integrates seamlessly with state-of-the-art grasp pose generation methods.
We validate the approach through large-scale simulations involving 15k grasps across 478 objects on three robotic hands, and real-world experiments on 8 objects. Results demonstrate that our method achieves higher grasp success rates and reduced undesired object movements. Supplementary materials are available at {\url{https://ada-grasp-ctrl.github.io/}}.
\end{abstract}

\section{Introduction}

Dexterous grasping is a fundamental skill in multi-fingered manipulation, as precise grasps often serve as prerequisites for subsequent tasks \cite{tang2024robotic,jiang2024contact,yang2025multi,yu2025robotic}. While substantial progress has been made in grasp pose synthesis and generation, comparatively less attention has been devoted to the actual execution of planned grasps, which demands delicate control of the physical hand-object interactions.

\begin{figure} [tb]
  \centering 
    \includegraphics[width=1.0\linewidth]{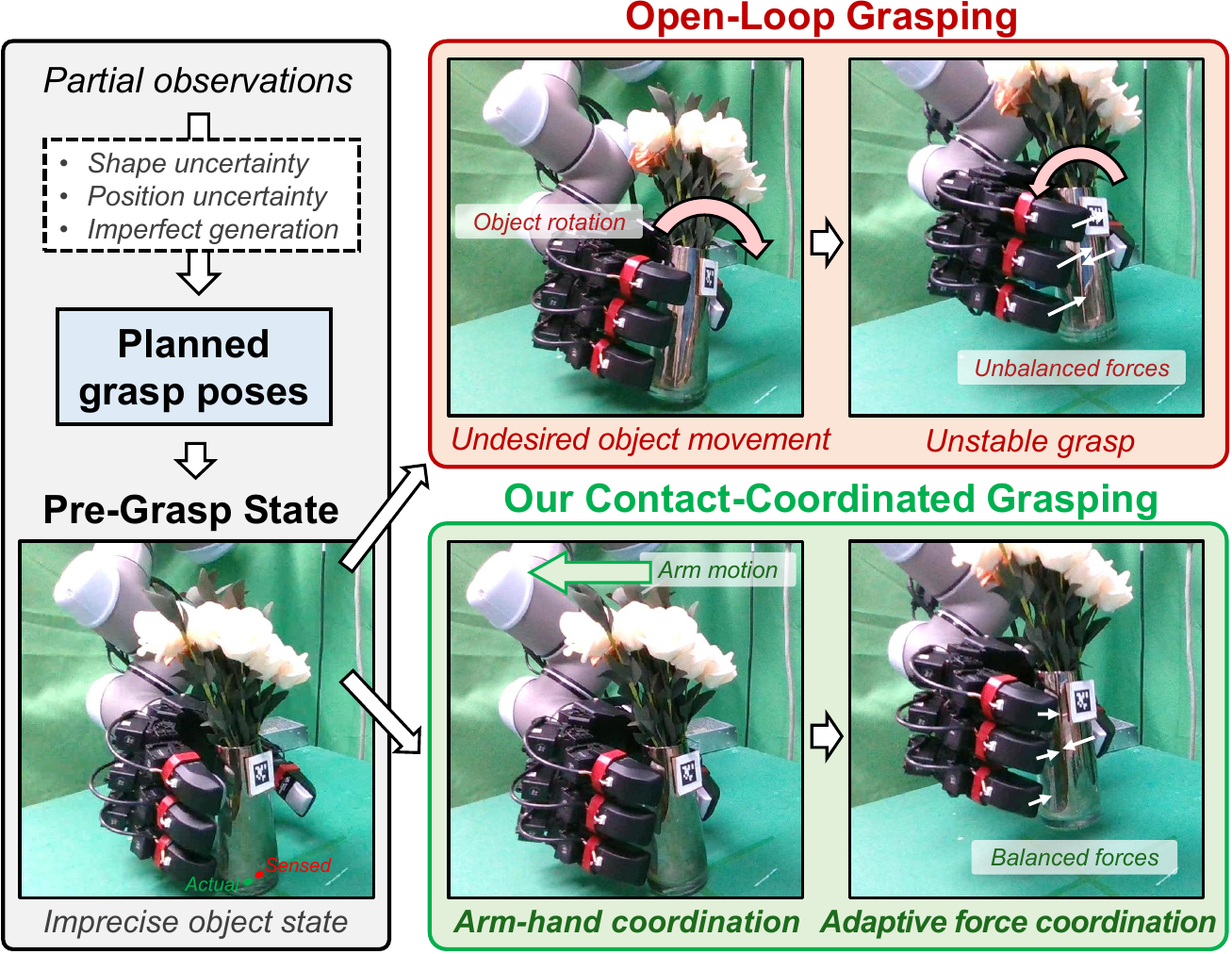} 
  \caption{\textbf{Execution of planned dexterous grasps.} 
  Due to uncertainty, open-loop execution of imperfect planned grasp poses can cause unintended in-hand object movements or grasp failures.
  This work proposes a tactile-driven control approach that coordinates multiple contacts during both approaching and grasping, reducing undesired object motion and enabling adaptive, delicate execution of diverse dexterous grasps.}
  \label{fig:fig1}
   \vspace{-5mm}
\end{figure}

We define a preferable \textit{delicate grasping process} as one where the robotic hand approaches an object on a horizontal tabletop and grasps the object without altering its original pose. Given a grasp pose planned by grasp pose generation approaches, the most straightforward and commonly used execution strategy is to move the robot to the planned joint configuration in an open-loop manner. However, the quality of generated grasps and open-loop execution can be affected by several practical factors, including 
1) \textbf{imperfect generation}: real-time grasp pose generation often relies on trained networks, which may produce small but impactful errors during inference;
2) \textbf{shape uncertainty}: partial observations and depth noises lead to uncertainty in the object's full geometry;
and 3) \textbf{position uncertainty}: errors in robot-camera calibration and sensing noises introduce uncertainty in the object's precise location.
Owing to perception and planning errors, we usually observe that objects are unintentionally moved in-hand during open-loop grasping, as multiple fingers fail to contact the object simultaneously or apply forces in a coordinated manner, as illustrated in Fig. \ref{fig:fig1}. These \textit{undesired object movements} may lead to incorrect grasp configurations or even grasp failures. 

This work aims to enable adaptive execution of planned tabletop grasps under uncertainty while minimizing undesired object movements through tactile-feedback control. 
The key for achieving this objective is maintaining \textbf{coordinated contacts} throughout both the approaching and grasping phases, meaning that contact forces from multiple fingers should remain balanced at all times. 
Existing tactile-feedback approaches typically treat approaching and grasping separately, primarily using position control during approaching to establish contacts and force control during grasping to apply desired forces \cite{takahashi2008adaptive,li2016dexterous,deng2020grasping,ford2023tactile,ford2025shear}. 
While this strategy can partly mitigate uncertainty, these methods usually control each finger independently and devote limited attention to the coordination among multiple fingers and contacts.

\begin{table*}
\centering
\caption{\textbf{Distinctive features of our approach compared with existing tactile-driven grasping control methods.}}
\vspace{-2mm}
\label{tab:method_comparison}
\begin{tabular}{cc|ccc|cc} 
\toprule
 Work & Method & \begin{tabular}[c]{@{}c@{}}Coordination-aware\\phase separation~\end{tabular} & \begin{tabular}[c]{@{}c@{}}Arm-hand \\coordination\end{tabular} & \begin{tabular}[c]{@{}c@{}}Adaptive force \\coordination\end{tabular} & \begin{tabular}[c]{@{}c@{}}Diverse \\planned grasps\end{tabular} & Evaluation scale \\ 
\hline
\cite{takahashi2008adaptive}  & Feedback control & \ding{55} & \ding{55} & \ding{55} & \ding{55} & 3 objects (real) \\
\cite{chen2015adaptive} & Feedback control & \ding{51} & \ding{55} & \ding{55} & \ding{55} & 12 objects (real) \\
\cite{li2016dexterous}  & Feedback control & \ding{55} & \ding{55} & \ding{55} & \ding{51} & 9 grasps, 3 objects (real) \\
\cite{deng2020grasping}  & Feedback control & \ding{55} & \ding{55} & \ding{55} & \ding{55} & 8 objects (real) \\
\cite{liu2022multi}  & Tactile servoing & \ding{55} & \ding{51} & \ding{55} & \ding{55} & 4 objects (sim) / 2 objects (real) \\
\cite{liang2021multifingered} & Reinforcement learning & - & \ding{55} & - & \ding{55} & 30 objects (sim) / 14 objects (real) \\
\cite{wang2022learning} & Imitation learning & - & \ding{55} & - & \ding{55} & 100 objects (real) \\
\cite{ford2023tactile} & Feedback control & \ding{55} & \ding{55} & \ding{55} & \ding{55} & 43 objects (real) \\
\cite{ford2025shear} & Feedback control & \ding{55} & \ding{55} & \ding{55} & \ding{55} & 3 objects (real) \\ 
\hline
Ours & MPC & \ding{51} & \ding{51} & \ding{51} & \ding{51} & \begin{tabular}[c]{@{}c@{}} 15k grasps, 478 objects, 3 hands (sim) \\8 grasps, 8 objects, 1 hand (real)~\end{tabular}\\
\bottomrule
\end{tabular}
\vspace{-5mm}
\end{table*}

To address the above objective, we propose a model predictive control (MPC) approach for coordinated contact control during grasping. The key contributions and novelties of our approach beyond existing methods include:
\begin{enumerate}
    \item \textbf{Coordination-Aware Phase Separation}: Unlike conventional methods that treat each finger independently, our approach separates the approaching and grasping phases based on the collective state of all contacts. The transition occurs once sufficient contacts are established to enable non-zero yet balanced forces.
    
    \item \textbf{Arm-Hand Coordination during Approaching}: The approaching phase is to establish adequate contacts on the object while avoiding large forces. To adapt to the actual object position without excessively deviating from the planned finger configuration, our method enables coordinated arm motions to adjust the global hand pose, contrasting with conventional methods that rely solely on finger motions.

    \item \textbf{Adaptive Force Coordination during Grasping}: The grasping phase is to coordinately increase contact forces to firmly grasp the object. Unlike conventional methods that prescribe fixed desired forces for each fingertip, our approach adaptively allocates forces of all contacts online based on measured contact locations and forces, guided by wrench balance criteria.
\end{enumerate}

To achieve these functions, our approach employs a contact-driven MPC with an analytical motion-contact model. 
Unlike methods constrained to fixed grasp types or fingertip-only contacts, our approach imposes no restrictions on the grasp postures, contact numbers, or locations, provided that tactile sensing of these contacts is available. 
Consequently, it can be seamlessly combined with state-of-the-art (SOTA) grasp pose generation methods and diverse grasp poses, enhancing actual grasp quality with tactile feedback as the only additional requirement. 

For evaluation, we perform a large-scale simulation study involving 15k grasps across 478 objects using three robotic hands, providing statistical validation of our approach on diverse grasps and comparisons with baselines. In addition, we demonstrate real-world deployment of the proposed method on eight everyday objects. The results show that our method achieves planned grasps with higher success rates and reduced undesired object motion under uncertainty.



\section{Related Work}

\subsection{Dexterous Grasp Planning}

Extensive research has been devoted to generating diverse multi-fingered grasp poses. Analytic methods typically compute force-closure grasps from complete object geometries, using either sampling-based search or gradient-based optimization with differentiable metrics, simulators, or bilevel formulations \cite{liu2021synthesizing,wang2023dexgraspnet,li2023gendexgrasp,chen2024bodex,chen2025dexonomy}. Building on large-scale synthesized datasets, learning-based approaches employ generative models for real-time grasp generation \cite{xu2023unidexgrasp,wei2024learning,wei2025d}, possibly with partial observations of objects. 
Since generated grasps are often imperfect, some open-loop strategies try to introduce intermediate poses (e.g., pre-grasp, grasp, squeeze) to improve robustness, reflecting coordination and phase separation from a purely kinematic perspective \cite{chen2024bodex,chen2025dexonomy}.

Some works considered object shape uncertainty in planning by placing contacts on object surface with lower uncertainty and applying higher finger compliance for contacts with higher uncertainty  \cite{chen2024springgrasp, li2016dexterous}. 
These approaches focus on improving the robustness of planned results with uncertainty modeled from visual observations. In contrast, our method focuses on closed-loop execution, leveraging real-time tactile feedback to adapt to actual positions and shapes.

\subsection{Tactile-Driven Dexterous Grasping Control} \label{sec:related_work}

Grasping control based solely on vision \cite{ wan2023unidexgrasp,qin2023dexpoint,liu2023dexrepnet} is challenging, as vision alone struggles to capture the precise hand-object relationship.
To address this, many works incorporated tactile sensing.
Some explored continuous re-grasping to adjust contact locations \cite{dang2013grasp,yan2024avita}
or resistance to disturbances after a stable grasp is achieved \cite{li2014learning,hang2016hierarchical,khadivar2023adaptive,winkelbauer2024learning}.
In contrast, our approach focuses on adaptive and delicate execution of diverse planned grasp, accounting for both the approaching and contact interactions.

\begin{figure*} [tb]
  \centering 
    \includegraphics[width=0.95\textwidth]{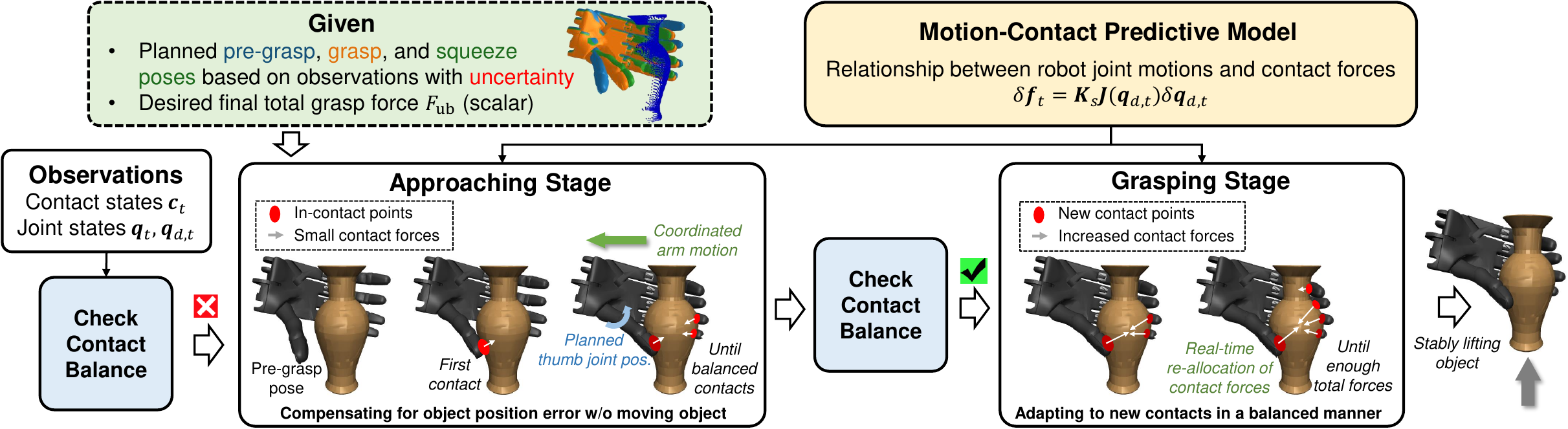} 
  \caption{\textbf{Overview of our tactile-driven coordinated contact control method} for adaptive execution of planned grasp poses generated from observations with uncertainty. Our method employs a coordination-aware separation of the approaching and grasping 
  phases, using the criteria of wrench balance. During the approaching phase, the fingers make contact with the object using gentle forces, while coordinated arm motions compensate for object position errors without deviating from the planned finger configurations. Once sufficient contacts are established, the fingers increase contact forces in a balanced manner to reach the desired total grasp force, during which the desired force of each contact is re-allocated in real time to adapt to changes in contact states.}
  \label{fig:overview}
   \vspace{-4mm}
\end{figure*}

A common tactile-driven grasping strategy is to use position control during approaching and force control after making contacts \cite{takahashi2008adaptive,li2016dexterous,deng2020grasping,ford2023tactile,ford2025shear}.
Tactile-servoing methods \cite{liu2022multi} map desired tactile changes to robot motions via tactile Jacobians.
Although these methods guarantee firm grasps of objects under shape uncertainty, they require pre-defined desired contacts for each fingertip and do not explicitly consider the multi-contact coordination, often leading to undesired object movements.
An approach-to-grasp strategy was proposed in \cite{chen2015adaptive} to handle position uncertainty, which pauses the first contacting finger until others make contacts. However, since only finger motions are used, the final contact configuration may deviate from the target grasp, resulting in unbalanced wrench and object rotation. 
Tactile-driven reinforcement learning \cite{liang2021multifingered} and imitation learning \cite{wang2022learning}, together with other based on reinforcement learning-based approaches \cite{zhang2024graspxl,zhang2025RobustDexGrasp}, improve grasp success but generally do not explicitly account for undesired object movements.
In contrast, our approach incorporates arm motions to compensate for object position errors and explicitly coordinates multiple contacts throughout both the approaching and grasping phases. Table \ref{tab:method_comparison} summarizes existing methods and emphasizes the distinguishing features of our approach.

\section{Preliminaries}


\textbf{Planned grasps}: 
The grasp pose generator we use plans three sequential grasp poses based on an point-cloud observation, including a collision-free pre-grasp pose, a grasp pose where the hand is just in contact with the object, and a squeeze pose that can firmly grasp the object, following \cite{chen2024bodex}.
We utilize these three poses as the guidance of finger moving directions in our control method.

\textbf{Assumptions}: 
We make the following assumptions for the mathematical formulation of the manipulation process: 1) All contacts between the hand and object are regarded as \textit{point contacts with friction} \cite{ferrari1992planning}; 2) The location, normal, and force of each contact can be measured by the tactile sensors in real time; 3) The fingers are driven by joint-space PD controllers; and 4) the robot is collision-free at the planned pre-grasp pose, and the fingers required for grasp are in contact with the object at the planned squeeze pose. 

\textbf{Notations}:
The joint position of the arm-hand robot at time $t$ is denoted as $\bm q_t \in \mathbb{R}^{N_a + N_h}$, where $N_a$ and $N_h$ are the degrees-of-freedom (DoFs) of the arm and hand, respectively. The contact state is denoted $\bm c_t = \{\bm c^{1}_t, \cdots, \bm c^{m}_t\}$, where $m$ is the number of contacts. The state of each contact $\bm c^{i}_t$ includes a contact force $\bm f^{i}_t \in \mathbb{R}^3$, a contact position  $\bm p^{i}_t \in \mathbb{R}^3$, and a contact normal $\bm n^{i}_t \in \mathbb{R}^3$. We denote $\bm f_t = [\bm f^{1}_t; \cdots; \bm f^{m}_t] \in \mathbb{R}^{3m}$. Note that we use $[\bm a; \bm b]$ to denote vertical concatenation of column vectors $\bm a$ and $\bm b$.

\textbf{Multi-contact grasp}:
Based on the contact normal $\bm n^{i}$, we can choose two tangent directions $\bm d^{i}$ and $\bm e^{i}$ to construct a contact frame $[\bm n^{i}, \bm d^{i}, \bm e^{i}]$.
Defining the contact force $\bm f^{i}$ in the contact frame,
the wrench $\bm w^i \in \mathbb{R}^6$ applied on the object from contact $i$ can be calculated as $\bm w^i = \bm G^i \bm f^i$, where 
\begin{equation}
    \bm {G}^i=
    \left[\begin{array}{c}
    \bm G^i_{F} \\
    \bm G^i_{M}
    \end{array}\right]
    =\left[\begin{array}{ccc}
    \bm {n}^i & \bm {d}^i & \bm {e}^i \\
    \tilde{\bm p}^i \times \bm {n}^i & \tilde{\bm p}^i \times \bm {d}^i & \tilde{\bm p}^i \times \bm {e}^i
    \end{array}\right] \in \mathbb{R}^{6 \times 3}
\end{equation}
We define $\tilde{\bm p}^i$ as the relative position to the centroid of all contact points. The total wrench is obtained as $\bm w = \sum_{i=1}^{m} \bm G^i \bm f^i = \bm G \bm f$, where $\bm G \in \mathbb{R}^{6 \times 3m}$. Additionally, we define a \textit{normalized wrench} as 
\begin{equation}
    \bar{\bm w} = \left[
    \frac{\bm G_{F} \bm f }{\sum_{i=1}^{m}\| \bm G^i_{F} \bm f^{i} \|_2}; 
     \frac{\bm G_{M} \bm f }{\sum_{i=1}^{m}\| \bm G^i_{M} \bm f^i \|_2} 
     \right]
\end{equation}

\section{Method}

\subsection{Overview of Contact-Coordinated Grasping Process} \label{sec:overview}

The best open-loop execution strategy leverages the planned three subsequent grasp poses.
The path from the pre-grasp pose to grasp pose aims to move from a collision-free pose to a pose just in contact (i.e., approaching phase), and the path from the grasp pose to squeeze pose aims to apply sufficient contact forces (i.e., grasping phase). 
This open-loop strategy has reflected the concept of coordination from solely kinematic perspective, as the fingers are always in coordinated configurations with respect to the object geometry.
However, since no feedback is utilized during execution, it cannot adapt to actual object positions and shapes in the presence of uncertainty and imperfect generations. Tactile feedback provides information about the actual contact states and has been widely used in parallel-gripper grasping to ensure firm grasps. However, how to effectively integrate it in multi-fingered dexterous grasping, so as to coordinate multiple contacts under uncertainty, remains an open question, which is the focus of our approach.

Our approach leverages the planned grasp poses as a guidance of finger moving directions. We first linearly interpolate between the pre-grasp, grasp, and squeeze pose, obtaining a joint-space \textit{guiding path} denoted as $\bar{\mathcal{P}} = \{ \bar{\bm q}_1, \cdots, \bar{\bm q}_T \}$. 
Due to uncertainty, the separation of the approaching and grasping phase 
should be adaptively determined based on real-time feedback.
Unlike most tactile-feedback methods that independently treat each finger, we incorporate a \textbf{coordination-aware separation} strategy, whose core criteria is whether the multiple in-contact locations can apply a balanced wrench to the object, as described in Section \ref{sec:phase_trans}. 

\begin{figure} [tb]
  \centering 
    \includegraphics[width=1.0\linewidth]{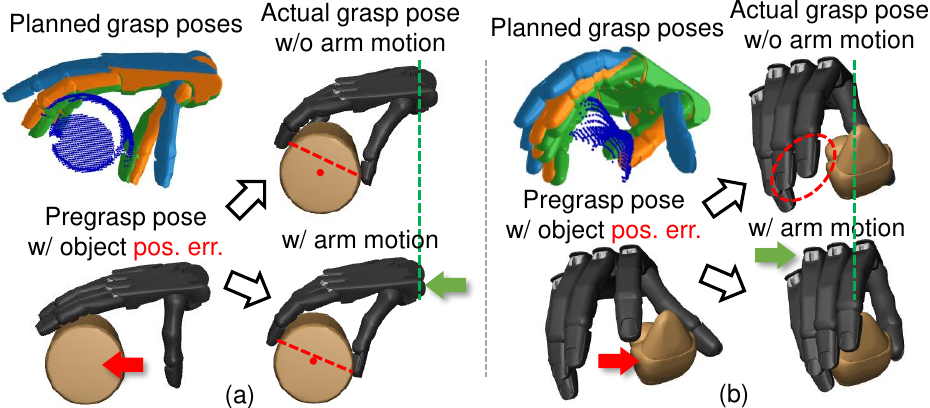} 
    \vspace{-5mm}
  \caption{\textbf{Two cases to illustrate the issues when only using finger motions to compensate for object position errors.} The baseline strategy is to stop the first-contacting fingers and actuate the remaining fingers until all fingers make contact, without involving arm motions (top-right figures of each case). It may lead to issues that (a) the fingers reach an unstable grasp configuration; or (b) the remaining fingers fail to make contact with the object even after reaching the planned final squeeze pose.}
  \label{fig:no_arm}
   \vspace{-5mm}
\end{figure}

During the first approaching phase, the objective is to move along the guiding path while avoiding applying large forces to the object, until satisfying the phase transition criteria. Owing to position uncertainty, the fingers may not make contact with the object simultaneously (as in Fig. \ref{fig:overview}). If the first in-contact fingers continue moving forward, they will unintendedly move the object. One strategy is to stop these fingers and actuate the other fingers until they are all in contact \cite{chen2015adaptive}. However, since the relative position between the object and hand palm deviates from the planned one, the first-contact fingers will undershoot the desired values while the other fingers will overshoot the desired values, leading to larger deviation from the planned contact configurations and less stable grasps, as illustrated in Fig. \ref{fig:no_arm}. 
Our key insight is that \textbf{arm-hand coordination} can be incorporated to simultaneously adjust the palm pose, enabling smaller deviation from the planned finger configurations. We design an MPC with joint optimization of arm and hand motions, 
which is 
described in Section \ref{sec:approaching}.

When the phase-transition criteria is satisfied, the robot starts to increase grasp forces. The key question is how to allocate the desired force of each contact. We use the wrench balance criteria to optimize the allocation of the force distribution across all contacts.
Note that, due to partial and noisy observations, it is uncertain prior to execution which parts of the hand will make contact with the object. Moreover, new contacts may still emerge after the phase transition (see Fig. \ref{fig:overview}).
Thus, the desired forces are re-allocated in real time to adapt to contact state changes, while also accounting for current contact forces and joint configurations to ensure smoothness and reachability.
To achieve this \textbf{adaptive force coordination}, we design an MPC that jointly optimizes the desired contact forces and corresponding robot motions.
as described in Section \ref{sec:grasping_phase}.

Furthermore, to control contact forces via robot joint movements, we employ a motion-contact model to predict contact force changes and incorporate it into the MPC for both the approaching and grasping phases, 
as described in Section \ref{sec:modeling}. The complete pipeline is illustrated in Fig. \ref{fig:overview}.

\begin{figure} [b]
    \vspace{-5mm}
  \centering 
    \includegraphics[width=0.45\linewidth]{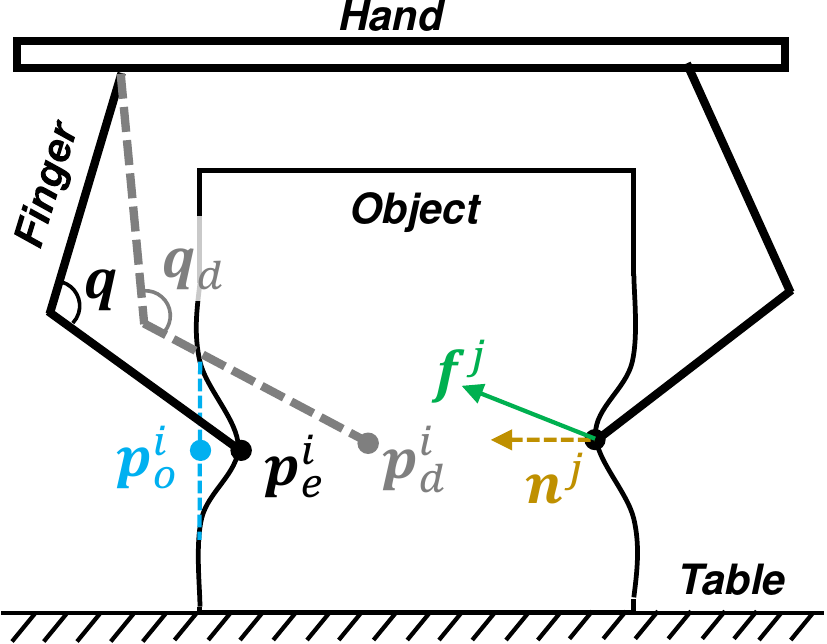} 
  \caption{\textbf{Illustration of the mathematical formulation of contacts.} The undeformed, actual, and desired positions of the $i^{\rm th}$ contact as well as the contact normals and forces of the $j^{\rm th}$ contact are shown.}
  \label{fig:notations}
\end{figure}

\subsection{Motion-Contact Modeling} \label{sec:modeling}

Inspired by the interaction control of robot arms \cite{gold2022model}, we employ a motion-contact model to predict the changes of contact forces caused by arm-hand joint movements.

Consider the $i^{\rm th}$ contact as illustrated in Fig. \ref{fig:notations}. Accounting for compliance of both the environment and robot joints, one contact involves three key positions: $\bm p_o^i, \bm p_e^i$ and $\bm p_d^i$. The $\bm p_o^i$ represents the undeformed position of the contact, and $\bm p_e^i$ represents the actual contact position. 
For convenience, all positions, forces, and Jacobians are expressed in local contact frames.
Under the quasi-static and linear-elastic assumption \cite{salisbury1980active}, the contact force $\bm f^i$ is determined by
\begin{equation} \label{eq:object_compliance}
    \bm f^{i} = \bm K_o \, \bm p_{o, e}^i = \bm K_o (\bm p_e^i - \bm p_o^i)
\end{equation}
where $\bm K_o$ is the object stiffness matrix.
The actual contact position $\bm p_e$ can be calculated using the robot forward kinematics as $\bm p_e^i = \text{FK}^i(\bm q)$. Due to the compliance of the robot joints, the \textit{actual} joint positions $\bm q$ will not be the same as the \textit{desired} joint positions $\bm q_d$ (i.e., the position control command). The aforementioned $\bm p_d^i$ represents the desired Cartesian-space position of $\bm q_d$, as $\bm p_d^i = \text{FK}^i(\bm q_d)$, and $\bm p_{e, d}^i = \bm p_d^i - \bm p_e^i$. 
Note that multiple contacts may exist on one finger. To consider their coupling, we stack all contact points as $\bm p = [\bm p^1_{(\cdot)}; \cdots; \bm p^m_{(\cdot)}]$ and Jacobians as $\bm J = [\bm J^1{(\cdot)}; \cdots; \bm J^m{(\cdot)}]$.
For quasi-static situations, we have
\begin{equation} \label{eq:joint_compliance}
    \bm J(\bm q)^\transpose \bm f = \bm \tau = \bm K_p (\bm q_d - \bm q)
\end{equation}
where $\bm \tau$ is joint torques and $\bm K_p$ is the position stiffness of the low-level joint PD controller. 
Considering that the object stiffness is usually much larger than the robot joint stiffness, the derivative ${\rm d} \bm q$ is usually much smaller than $ {\rm d} \bm q_d$ during in-contact movements. Thus, we approximate that ${\rm d}\bm p_{e, d} \approx \bm J(\bm q_d) ({\rm d} \bm q_d -  {\rm d} \bm q)$, instead of using $\bm J(\bm q)$ in \cite{gold2022model}.
Substituting it into the derivative of (\ref{eq:joint_compliance}) and neglecting the derivative of the Jacobian under quasi-static assumption, we have
\begin{equation}  \label{eq:joint_compliance_2}
    {\rm d}\bm p_{e, d} = \bm J(\bm q_d) \bm K_p^{-1} \bm J(\bm q)^\transpose \, {\rm d}\bm f 
\end{equation}
Then, substituting (\ref{eq:joint_compliance_2}) into the derivative of (\ref{eq:object_compliance}), we have
\begin{equation}
\begin{aligned}
     {\rm d}\bm f &= \bar{\bm K_o} \, {\rm d}\bm p_{o, e} = \bar{\bm K_o} {\rm d}\bm p_{o, d} - \bar{\bm K_o} {\rm d}\bm p_{e, d} \\
     &=  \bar{\bm K_o} {\rm d}\bm p_{o, d} - \bar{\bm K_o} \bm J(\bm q_d) \bm K_p^{-1} \bm J(\bm q)^\transpose \, {\rm d}\bm f 
\end{aligned}
\end{equation}
\begin{equation} \label{eq:contact_model}
\begin{aligned}
    {\rm d}\bm f = \bm K_s {\rm d}\bm p_{o, d} = \left( \bm I + \bar{\bm K_o} \bm J(\bm q_d) \bm K_p^{-1} \bm J(\bm q)^\transpose \right)^{-1}\bar{\bm K_o} {\rm d}\bm p_{o, d}
\end{aligned}
\end{equation}
where $\bar{\bm K_o}$ is the diagonal stack of ${\bm K_o}$.
If assuming a temporarily constant object pose (i.e., ${\rm d}\bm p_o = \bm 0$), we have
\begin{equation} \label{eq:single_contact_model}
    {\rm d}\bm f = \bm K_s \bm J(\bm q_d) \, {\rm d}\bm q_d
\end{equation}
Note that $\bm K_s \rightarrow \left(\bm J(\bm q_d) \bm K_p^{-1} \bm J(\bm q)^\transpose \right)^{-1}$ when $\bm K_o \rightarrow \infty$.
Thus, $\bm K_o$ also acts like a damped regularization term to tackle singularities of the finger Jacobians. Additionally, as assuming temporarily constant object poses, the (\ref{eq:single_contact_model}) ignores the potential coupling among multiple contacts caused by slight object movements. The full coupling effects are influenced by frictional and supporting forces/torques from the table, which are difficult to model precisely in practice.


\subsection{Model Predictive Grasping Control} \label{sec:mpgc}


The optimization variables for both phases include control input $\bm u_t = \bm q_{d, t+1} - \bm q_{d, t}$ and desired forces of all contact $\bm f_{t+1}$. 
Note that this formulation does not take account of the future contact states of current non-contacting points, as they cannot be predicted using tactile feedback alone.

\subsubsection{Approaching Phase} \label{sec:approaching}

The MPC  
is formulated as
\begin{align}
    \min_{\bm u_t, \bm f_{t+1}} \quad & 
    \mathcal{J}_{\text{guide}} + \mathcal{J}_{\text{palm}} + \mathcal{J}_{\text{tm}} + \mathcal{J}_{\text{vel}} + \mathcal{J}_{\text{acc}} \\
    \text{s.t.} \quad 
    & \quad \bm f_{i,t+1} = \bm f_{i,t} + \bm K_s \bm J(\bm q_{d,t}) \bm u_t \\
    & \quad \sum_{i=1}^{m} f_{i,1,t+1} \leq F_{\text{appr}} \label{eq:cm_constraint} \\
    & \quad \bm q_{\text{lb}} \leq \bm q_{d, t} + \bm u_t \leq \bm q_{\text{ub}}
\end{align}
Constraint (\ref{eq:cm_constraint}) ensures that the fingers do not apply large forces to the object, where $f_{i,1,t+1}$ represents the normal-axis force and $F_{\text{appr}}$ is a small threshold. 
The cost $\mathcal{J}_{\text{guide}}$ aims to follow the planned guiding path $\bar{\bm P}$ of fingers as 
$\mathcal{J}_{\text{guide}} = \| (\bm q_{d,t} + \bm u_t) - \bar{\bm q}_{t+1} \|_{\bm W_g}$,
where the elements for arm joints in the weight matrix $\bm W_g$ are zeros.
The cost $\mathcal{J}_{\text{palm}}$ aims to  restrict the vertical-axis movement and rotation of the hand palm, specified as $\mathcal{J}_{\text{palm}} = \| \text{FK}^{p}(\bm q_{d,t} + \bm u_t) - \text{FK}^{p}(\bar{\bm q}_{t+1}) \|_{\bm W_p}$,
where the horizontal axes of the weight matrix $\bm W_p$ are zeros. 
The cost $\mathcal{J}_{\text{tm}}$ penalizes the tangential movements of the contact points, specified as $\mathcal{J}_{\text{tm}} = \| \bm J(\bm q_{d,t}) \bm u_t \|_{\bm W_t}$,
where the tangential dimensions of $\bm W_t$ are non-zero. Additionally, the cost $\mathcal{J}_{\text{vel}}$ and $\mathcal{J}_{\text{acc}}$ slightly penalize the velocity and acceleration of $\bm q_d$ to enhance smoothness.

This formulation enables coordinated arm motions to adjust the palm position and adapt to the actual object position, avoiding large contact forces while following the guiding finger path as close as possible.

\begin{figure*} [tb]
  \centering 
    \includegraphics[width=1.0\textwidth]{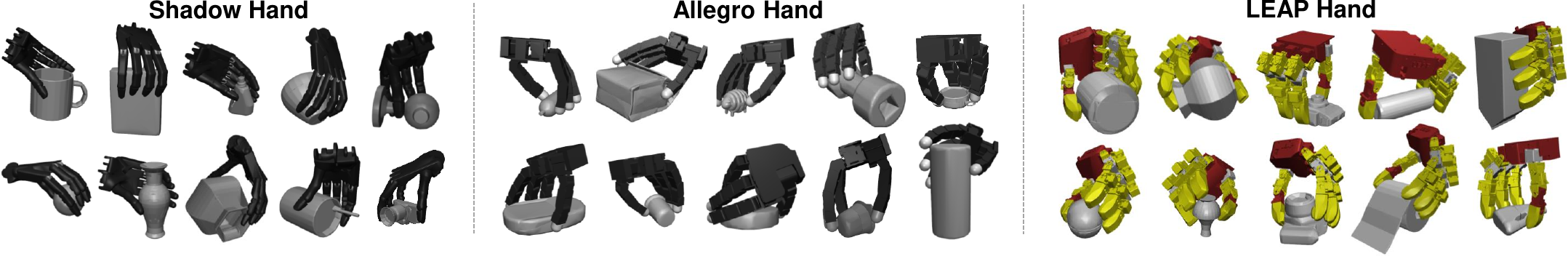} 
  \vspace{-7mm}
  \caption{\textbf{Visualization of some grasps in our large-scale simulation evaluation.} The evaluation totally involves 15k grasps on 478 objects and three dexterous hands. These figures show the grasp states after lifting from a table, achieved using our approach.}
  \label{fig:diverse_grasps}
\end{figure*}

\begin{table*}
\centering
\setlength{\tabcolsep}{5.5pt} 
\caption{\textbf{Large-scale simulation evaluation results under object shape uncertainty.}}
\vspace{-2mm}
\label{tab:sim_compare}
\begin{threeparttable}[b]
\begin{tabular}{l|cccc|cccc|cccc} 
\toprule
\multirow{2}{*}{Method} & \multicolumn{4}{c|}{Shadow} & \multicolumn{4}{c|}{Allegro} & \multicolumn{4}{c}{LEAP} \\
 & SR\tnote{a} $\uparrow$ & Pos. $\downarrow$ & Rot. $\downarrow$ & Wrench $\downarrow$ & SR $\uparrow$ & Pos. $\downarrow$ & Rot. $\downarrow$ & Wrench $\downarrow$ & SR $\uparrow$ & Pos. $\downarrow$ & Rot. $\downarrow$ & Wrench $\downarrow$ \\ 
\hline
Open-loop        & 81.6 & 6.0\std{5.5} & 5.2\std{4.6} & 0.53\std{0.24} & 79.8 & 5.9\std{5.5} & 5.0\std{4.1} & 0.44\std{0.19} & 79.3 & 7.3\std{5.5} & 5.3\std{4.3} & 0.43\std{0.20} \\
Feedback control & 86.7 & 18\std{11}   & 13\std{12}   & 0.43\std{0.19} & 86.1 & 18\std{11}   & 12\std{9.8}  & 0.44\std{0.17} & 87.8 & 20\std{13}   & 11\std{8.5}  & 0.45\std{0.20} \\
W/o arm motion   & 90.8 & 3.1\std{5.2} & 3.4\std{4.9} & 0.18\std{0.19} & 91.5 & 2.8\std{4.7} & 2.9\std{4.2} & 0.19\std{0.17} & 93.2 & 3.2\std{5.3} & 3.3\std{4.5} & 0.18\std{0.18} \\
Independ. forces & 87.7 & 12\std{6.5}   & 7.9\std{6.4} & 0.42\std{0.18} & 86.2 & 12\std{6.6}   & 7.4\std{5.6}  & 0.43\std{0.15} & 89.2 & 14\std{9.1}   & 7.0\std{5.2}  & 0.46\std{0.19} \\ 
\hline
\textbf{Ours} & \textbf{91.6} & \textbf{2.4\std{4.2}} & \textbf{2.6\std{4.5}} & \textbf{0.16\std{0.17}} & \textbf{92.4} & \textbf{2.2\std{4.3}} & \textbf{2.2\std{3.7}} & \textbf{0.18\std{1.15}} & \textbf{93.6} & \textbf{2.5\std{4.7}} & \textbf{2.4\std{3.5}} & \textbf{0.17\std{0.17}} \\
\bottomrule
\end{tabular}
\begin{tablenotes}
      \item[a] SR: success rate (\%). Pos. (mm): averaged object position errors. Rot. ($^{\circ}$): averaged object rotation error. Wrench: averaged normalized wrench. Results are reported as (mean \std{standard deviation}).
\end{tablenotes}
\end{threeparttable}
\end{table*}

\begin{figure*} [tb]
  \centering 
    \includegraphics[width=1.0\linewidth]{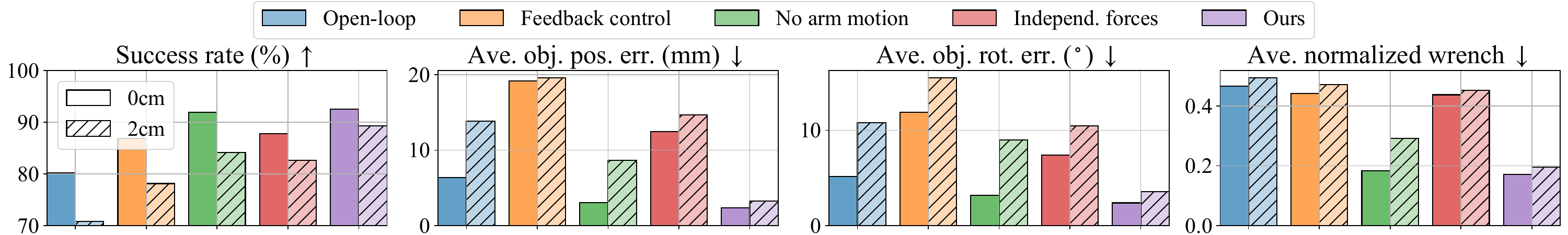} 
   \vspace{-5mm}
  \caption{\textbf{Simulation evaluation results under object position uncertainty}, averaged across the three hands. ``0cm" refers to no position errors, and ``2cm" refers to perturbing the positions by 2 cm along eight uniformly distributed planar directions respectively.}
  \label{fig:exp:sim_pos_err}
   \vspace{-5mm}
\end{figure*}

\subsubsection{Grasping Phase} \label{sec:grasping_phase}
The MPC is formulated as
\begin{align}
    \min_{\bm u_t, \bm f_{t+1}} \quad & 
    \mathcal{J}_{\text{guide}} + \mathcal{J}_{\text{wrench}} + \mathcal{J}_{\text{vel}} + \mathcal{J}_{\text{acc}} \\
    \text{s.t.} \quad 
    & \quad \bm f_{i,t+1} = \bm f_{i,t} +  \bm K_s \bm J(\bm q_{d,t})\bm u_t \\
    & \quad \sum_{i=1}^{m} f_{i,1,t+1} = F_{\text{des}, t+1} \label{eq:cm_constraint_2} \\
    & \quad \bm f_{i,t+1} \in \mathcal{F}, \quad i \in [1, m] \label{eq:friction_cone}  \\
    & \quad \bm q_{\text{lb}} \leq \bm q_{d, t} + \bm u_t \leq \bm q_{\text{ub}} \\
    & \quad  \bm W_a \bm u_t = \bm 0 \label{eq:arm_constraint}
\end{align}
Constraint (\ref{eq:cm_constraint_2}) ensures that the sum of normal contact forces equals a scalar desired grasp force $F_{\text{des}, t}$. We define it as an increased value over time $t$:
\begin{equation}
    F_{\text{des}, t+1} = \min \left (\max \left(\sum_{i=1}^{m} f_{i,1,t}, F_{\text{des}, t} \right), F_{\text{ub}} \right) + \delta F
\end{equation}
where $F_{\text{ub}}$ is the pre-defined final grasp force and $\delta F$ is the step size.
Constraint (\ref{eq:friction_cone}) represents the friction cone constraint formulated as 
$f_{i,1} \geq 0$ and $\sqrt{f_{i,2}^2 + f_{i,3}^2} \leq \hat{\mu}\, |f_{i,1}|$
for $i \in [1, m]$, where $\hat{\mu}$ is the estimated friction coefficient. It is usually safe to set $\hat{\mu}$ below the actual value in practice. 
The cost $\mathcal{J}_{\text{wrench}}$ is to adaptively coordinate desired forces of multiple contacts by pursuing wrench balance, which is specified as $\mathcal{J}_{\text{wrench}} = \| \bm G \bm f_{t+1} \|_2^2$.
The $\mathcal{J}_{\text{guide}}$ is to actuate the non-contact fingers to continue following the guiding path to establish potential additional contacts, and the element of its weight matrix $\bm W_g$ is set to zero if the corresponding finger is already in contact. Additionally, arm motions are not used in this phase and constrained by (\ref{eq:arm_constraint}), where the arm-related element of $\bm W_a$ are set to 1.

Note that this formulation requires a pre-defined final grasp force $F_{\text{ub}}$ which ensures a firm grasp. The work does not tackle the problem of determining $F_{\text{ub}}$, as it depends on the object's mass, friction coefficient, and material softness, which cannot be fully addressed from a control perspective. Possible solutions include visual estimation of object properties via vision-language models or tactile-based slippage detection to increase $F_{\text{ub}}$ reactively \cite{su2015force,deng2020grasping}.

\subsubsection{Phase Transition} \label{sec:phase_trans}

The criteria of phase transition is defined as whether the multiple in-contact locations can apply a balanced wrench to the object. It is online calculated via the following optimization problem:
\begin{align}
    &  \min_{\bm f}  
      \quad J = \| \bm G \bm f \|_2^2 \\
    \text{s.t.} \quad 
    & \bm f_{i} \in \mathcal{F}, \, i \in [1, m] \quad \text{and} \quad \sum_{i=1}^{m} f_{i,1} = 1
\end{align}
 Considering object scales, we express torques in units of $\mathrm{N{\cdot}cm}$ when computing $J$. 
The grasping phase is activated if the normalized wrench $\| \bar{\bm w} \|_2$ of the optimal solution is below a threshold $\epsilon_b$. 






\section{Results}

In the experiments, the guiding paths are constructed by linearly interpolating between planned pre-grasp and grasp poses over 2 seconds, and between grasp and squeeze poses over another 2 seconds. The action is computed at 10Hz and then interpolated for high-frequency low-level execution. 
The hyper-parameters and more implementation details are provided in Appendix (available on \href{https://ada-grasp-ctrl.github.io/}{Website}).
The optimization is solved using \texttt{scipy.optimize.minimize}  in Python.

\textbf{Evaluation metrics}:
The metrics for evaluation includes 1) \textbf{Success rate} (\%): a grasp is considered successful if the object is lifted from the tabletop and reaches the desired height;
2) \textbf{Average object position error} (mm): the distance between the final object position and desired position (i.e., initial object position + lift height), averaged over all successful cases;
3) \textbf{Average object rotation error} ($^{\circ}$): the angle between the final and initial object orientation, averaged over all successful cases;
and 4) \textbf{Average normalized wrench}: the normalized wrench during the last 0.5 second before lifting, averaged over all success cases.

\textbf{Baselines}:
We choose four baselines for comparison in simulation, including 1) \textbf{Open-loop}: The open-loop execution strategy as aforementioned.
2) \textbf{Feedback control}: A widely used two-phase tactile-feedback controller concluded from \cite{takahashi2008adaptive,li2016dexterous,deng2020grasping,ford2023tactile,ford2025shear}, where each finger are independently controlled. It uses position control for the approaching phase and force control after making contacts. Here the force control law is designed as $\delta \bm q^i_d = \alpha \bm (\bm K_p^{i})^{-1} \bm J^i(\bm q)^\transpose (\bm f^i_{d} - \bm f^i)$, where the desired force $\bm f^i_{d}$ are set to be equal for all contacts.
3) \textbf{Without arm motion}: Only adjusting finger motions to adapt to object position errors without leveraging arm motions, like \cite{chen2015adaptive}. We implement it by forbidding arm motions in our framework. 
4) \textbf{Independent forces}: Using independent equal desired forces for each contact in the grasping phase, while other settings are the same as ours. 
The first two baselines are well-practiced strategies, and the last two are ablations of ours to respectively show the significance of coordinated arm motions and adaptive contact force coordination.
For fair comparison, the desired total force $F_{\rm ub}$ in our method is set to match the averaged final force observed in open-loop execution, and the desired force of each contact in baselines (2) and (4) is set to $F_{\rm ub} / N_{\rm finger}$ (the number of fingers).

\textbf{Grasp pose planning}:
As for the grasp pose generation, we first adopt one of the SOTA grasp synthesis approaches \cite{chen2024bodex} to construct a large-scale tabletop grasp pose dataset for the Shadow, Allegro, and LEAP Hand, respectively. To achieve real-time grasp inference from partial observations, we then train flow-based generative models as in \cite{chen2024bodex,chen2025dexonomy} using the synthesized grasp poses and single-view point clouds of objects. These models have demonstrated reliable performance and are well applied in the real world. Specifically, 2k objects from the DGN assets \cite{wang2023dexgraspnet} are divided into training and testing sets with a 4:1 ratio. Each object is scaled to multiple sizes, placed on the tabletop in different poses, and rendered from various single-view perspectives. The trained models are employed for grasp pose planning.

\begin{figure*} [tb]
  \centering 
    \includegraphics[width=1.0\linewidth]{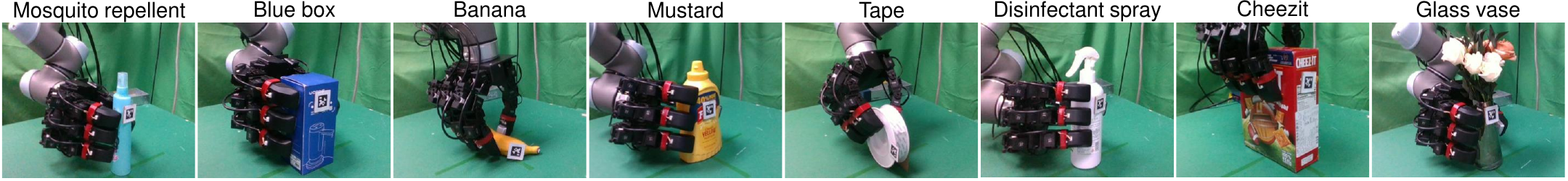} 
  \vspace{-5mm}
  \caption{\textbf{Real-world experiments include comparison of our method with baselines on eight different everyday objects and planned grasps.} The AprilTag markers are only used for quantitative evaluation of undesired object movements.}
  \label{fig:exp:real_objects}
   \vspace{-2mm}
\end{figure*}

\begin{table*}[t]
\centering
\caption{\textbf{Quantitative results of the real-world experiments}, averaged across all tests.}
\vspace{-2mm}
\label{tab:exp:real_compare}
\begin{threeparttable}[b]
\begin{tabular*}{\textwidth}{@{\extracolsep{\fill}} l|ccccc|ccccc} 
\toprule
\multirow{2}{*}{Method} & \multicolumn{5}{c|}{Shape uncertainty (8 objects)} & \multicolumn{5}{c}{Position uncertainty (2 objects, 4 positions)} \\
 & SR\tnote{a}  $\uparrow$ & Pos. $\downarrow$ & Rot. 1 $\downarrow$ & Rot. 2 $\downarrow$ & Wrench $\downarrow$ & SR $\uparrow$ & Pos. $\downarrow$ & Rot. 1 $\downarrow$ & Rot 2 $\downarrow$ & Wrench $\downarrow$ \\ 
\hline
Open-loop        & 40/40 & 8.5\std{4.6} & 3.5\std{2.2} & 6.7\std{2.5} & 0.57\std{0.19} & 20/20 & 30\std{13}  & 12\std{4.1}  & 12\std{5.6}  & 0.74\std{0.09} \\
Feedback control & 40/40 & 11\std{6.6}  & 3.4\std{2.1} & 5.3\std{3.2} & 0.54\std{0.15} & 20/20 & 13\std{7.4}  & 4.9\std{1.7} & 8.5\std{4.9} & 0.69\std{0.13} \\ 
\hline
\textbf{Ours} & \textbf{40/40} & \textbf{3.1\std{2.4}} & \textbf{1.7\std{1.1}} & \textbf{3.1\std{1.4}} & \textbf{0.40\std{0.12}} & \textbf{20/20} & \textbf{4.1\std{1.6}} & \textbf{3.2\std{1.7}} & \textbf{5.2\std{1.9}} & \textbf{0.42\std{0.09}} \\
\bottomrule
\end{tabular*}
\begin{tablenotes}
      \item[a] SR: success rate. Pos. (mm): object position error before lifting. Rot. 1 ($^{\circ}$): object rotation error before lifting. Rot. 2 ($^{\circ}$): object rotation error after lifting. Wrench: normalized wrench before lifting. Results are reported as (mean \std{standard deviation}).
\end{tablenotes}
\end{threeparttable}
\vspace{-7mm}
\end{table*}

\begin{figure} [tb]
  \centering 
    \includegraphics[width=1.0\linewidth]{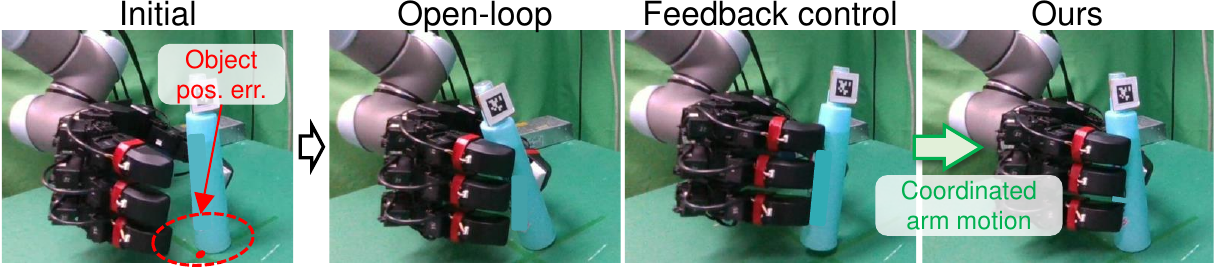} 
  \caption{\textbf{Comparison between the methods under position uncertainty.}}
  \label{fig:exp:real_pos_err}
   \vspace{-5mm}
\end{figure}

\subsection{Simulation Studies}

We conduct simulation studies to statistically evaluate the methods. 
We run the trained models on the test set and select top 10 grasps from 100 batched samples for each case based on probability estimation. We then randomly sample 5k grasps across 478 objects for each hand, which are used in the following evaluation. 
The grasping is simulated in MuJoCo \cite{todorov2012mujoco}. Some of the grasps are visualized in Fig. \ref{fig:diverse_grasps}.

\subsubsection{Evaluation with Shape Uncertainty} \label{sec:sim_shape_uncertainty}

Since the network-based generation uses partial pointcloud as the observation, the generated grasps reflect the influence of shape uncertainty. The evaluation results on all grasps of the three hands using our method and the baselines are summarized in Table \ref{tab:sim_compare}. The results indicate that 1) our approach performs similarly to the baseline without arm motion, as the visual observations contain no position errors; 2) our approach performs consistently better than other baselines on all three hands, demonstrating the significance of balanced grasping forces; 3) tactile-feedback approaches achieve higher success rates than open-loop execution, as they try to guarantee sufficient forces in every case; and 4) unbalanced forces from multiple contacts lead to large undesired object movements, even worse than open-loop executions that preserves kinematic-level finger coordination.

\subsubsection{Evaluation with Position Uncertainty}

We further evaluate the methods under object position errors. Specifically, the initial object positions are perturbed by 2 cm along eight uniformly distributed planar directions respectively to mimic position uncertainty. Cases with in-collision pre-grasp poses are filtered out. The averaged results across the three hands are shown in Fig. \ref{fig:exp:sim_pos_err}. The results indicate that 1) our method achieves the best performance, with the highest success rate as well as the lowest object movements and wrenches; 2) our method exhibits the smallest performance degradation between the settings with and without perturbation, demonstrating its higher adaptability to position errors; and 3) the baseline without arm motions suffers from higher performance drops, highlighting the significance of incorporating coordinated arm motions during approaching.

\subsection{Real-World Experiments}

Real-world experiments are conducted on a UR5 arm and a LEAP Hand \cite{shaw2023leaphand}. Each fingertip is equipped with a vision-based tactile sensor named Tac3D \cite{zhang2022tac3d}, which can robustly estimate the contact surface shape and three-axis contact force distribution. Single-view point clouds of objects are captured by a calibrated Azure Kinect DK camera for grasp generation. To quantitatively evaluate the object movements during grasping, an AprilTag marker is attached to each object and tracked by another RealSense camera. Since the Realsense is not extrinsically calibrated, we report the object position and rotation errors before lifting and the rotation errors after lifting, all computed as relative movements with respect to the initial pose. 
As tactile sensing is currently available only at the fingertip pads, we restrict our real-world experiments to fingertip grasps. However, our formulation imposes no constraints on contact distribution, as demonstrated in the simulation evaluation. The time cost of MPC solving is around 30 ms on an i9-13900K CPU.

\subsubsection{Evaluation with Shape Uncertainty}
We evaluate our approach against the open-loop strategy and feedback control method on eight planned grasps of different objects, as shown in Fig. \ref{fig:exp:real_objects}. Each grasp is executed five times, and the averaged results across all objects are reported in Table \ref{tab:exp:real_compare} (left). The results show that our approach effectively reduces the undesired object movements.

\subsubsection{Evaluation with Position Uncertainty}
We further conduct experiments under large position errors, using the glass vase and mosquito repellent bottle. The object is displaced by approximately 2 cm from its original position, either towards the thumb or index finger. Each case is executed five times, and the averaged results are reported in Table \ref{tab:exp:real_compare} (right). An example is shown in Fig. \ref{fig:exp:real_pos_err}, which shows that our method successfully adapts to the actual object position by leveraging coordinated arm motions, whereas the baselines lead to significant undesired object movements. 

\section{Discussion and Conclusion}

\textbf{Discussion}:
We highlight aspects that are not yet fully explored and potential future improvements:
1) The computing efficiency and control frequency have not yet been fully optimized. The optimization could be solved faster with more efficient solvers. We expect that higher control frequency would further improve force control performance.
2) In the real-world experiments, tracking dynamic desired forces is not very accurate, especially the tangential forces, likely due to limitations in hand's actuation mechanism, tactile sensors, and control delays. Despite these practical challenges, our method still outperforms the baselines.
3) Control of the lifting phase will be studied in future work. For instance, $F_{\rm ub}$ could be adjusted online using slippage detection during lifting, and the direction of frictions could be actively controlled to counteract the gravity.

\textbf{Conclusion}:
This work proposes a tactile-driven MPC for coordinated contact control that enables adaptive and delicate execution of planned dexterous grasps under uncertainty. By maintaining coordination across multiple contacts during both the approaching and grasping phases, our approach minimizes undesired in-hand object movements and improves grasp qualities. It is compatible with diverse grasp poses planned by existing well-developed grasp pose generation methods. We evaluate our approach in simulation with 15k grasps across 478 objects on three robotic hands, and in real-world experiments on 8 objects. The results demonstrate that our method achieves planned grasps under uncertainty with higher success rates and reduced undesired object movements compared with the baselines.









{
\small
\bibliographystyle{IEEEtran}
\bibliography{ref}
}

{\appendices

\section{Implementation Details}

\subsection{Identification of Robot Joint Stiffness}

In the simulation, we use the ground truth values of the robot joint stiffness $\bm K_p$, which are specified in the MuJoCo XML files.

In the real world, we first identify $\bm K_p$ of the fingers. The identification method is as follows. Under quasi-static conditions, we have
\begin{equation}
    \bm J(\bm q)^\transpose \bm f = \bm K_p (\bm q_d - \bm q)
\end{equation}
We command one finger to press its fingertip against a fixed object while gradually increasing the force. During this process, we collect the sensed forces $\bm f^i$, joint positions $\bm q$, and commanded positions $\bm q_d$. Then, we apply the linear-squares method to estimate $\bm K_p$ from the collected data. Since the LEAP Hand uses identical motors for all joints, we assume that $k_p$ of all joints are the same. 
The estimated $k_p$ is approximately 0.8.

It should be noted that the Dynamixel XC-330-M288-T motors on the LEAP Hand actually use a low-level current-based PID controller for high-level position control, and the relationship between joint torques and currents is slightly nonlinear 
(\href{https://emanual.robotis.com/docs/en/dxl/x/xc330-m288/}{\text{url}}
\footnote{
\href{https://emanual.robotis.com/docs/en/dxl/x/xc330-m288/}{https://emanual.robotis.com/docs/en/dxl/x/xc330-m288/}
}). 
Therefore, the identified $\bm K_p$ should be regarded as an approximation of the actual motor mechanism.

\subsection{Hyper-Parameters}

The hyper-parameters are set as $F_{\rm appr} = 0.2$N, $\epsilon_b=0.2$, $\hat{\mu}=0.3 (\text{sim}) / 0.7(\text{real})$, $\bm W_q = 1.0$ for finger joints and $0$ for arm joints, $\bm W_p = \text{diag}(0, 0, 10^{2}, 10, 10, 10)$, $\bm W_t = \text{diag}(0, 10^{3}, 10^{3})$, and $\bm K_o = 10^{5}\times \bm I_3 (\text{sim}) / 10^{4}\times \bm I_3(\text{real})$.
The $\delta F$ is set as $\delta F = F_{\text{ub}} / L_g$, in which $L_g$ is the number of waypoints in the interpolated guiding path from the grasp pose to squeeze pose. 
These values are straightforward to select manually based on reasonable estimates and have not been extensively tuned.


\subsection{Pre-Grasp Poses}

To obtain a pre-grasp pose with larger error tolerance to reduce the risk of initial collision under object position uncertainty, we re-compute the pre-grasp pose as the linear extrapolation from the grasp pose towards the original pre-grasp pose generated by the network (three times the original distance).

\subsection{Real-world Tactile Sensing}

We use Tac3D, a vision-based tactile sensor, to measure contact states in the real world, as shown in Fig. \ref{fig:hardware}. By visually tracking markers on the sensor surface, the sensor estimates local surface deformations and, based on factory calibration, the 3-axis contact forces of each taxel. 
The total contact force $\bm f^i$ is directly provided by the sensor.
To obtain the contact position, we segment the in-contact region using a force threshold and take its centroid. 
The contact normal $\bm n^i$ is then estimated from the surface normal at the contact position, and the global contact position $\bm p^i$ is obtained through the robot’s forward kinematics. 
We experimentally find that the sensed normal force is relatively accurate, with relative errors below 10\%. The directions of the sensed tangential forces are accurate, and their magnitudes appear reasonable, although we have not quantitatively evaluated them. Note that the valid region for reliable tactile sensing is approximately a 2 cm $\times$ 2 cm plane, which limits the application of our approach in current real-world experiments.

\begin{figure} [tb]
  \centering 
    \includegraphics[width=0.7\linewidth]{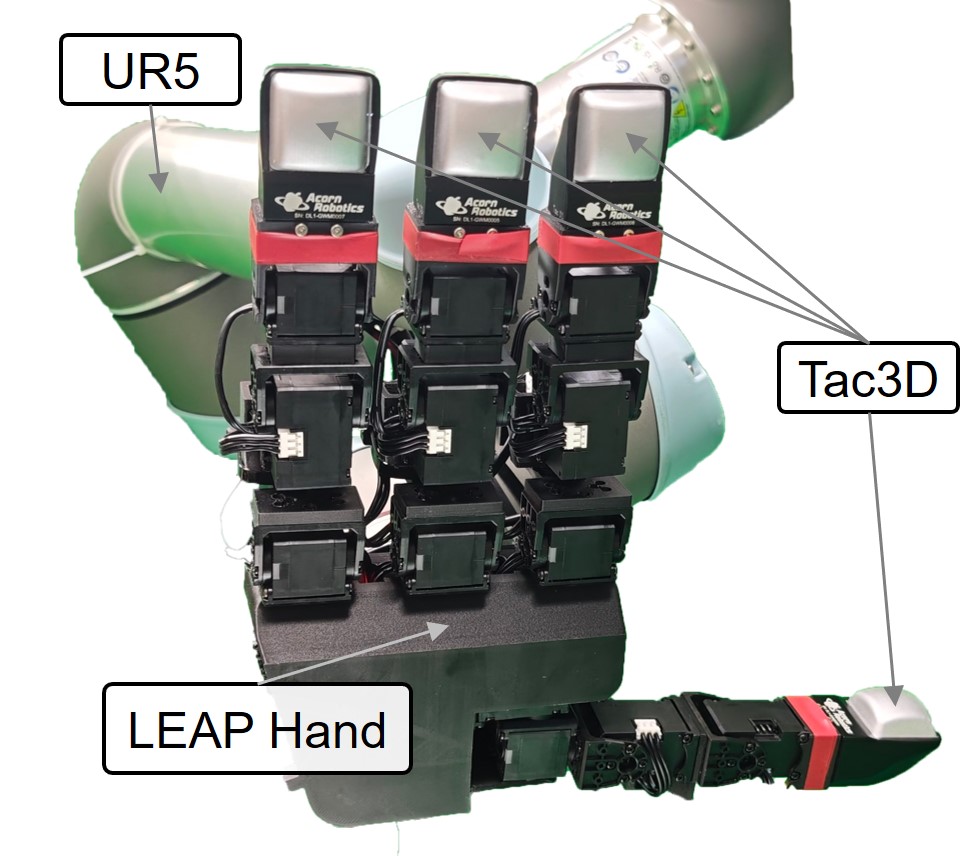} 
  \caption{\textbf{Hardware setup for real-world experiments.} We use Tac3D, a vision-based tactile sensor, to measure the contact states.}
  \label{fig:hardware}
\end{figure}

\begin{figure} [b]
  \centering 
    \includegraphics[width=1.0\linewidth]{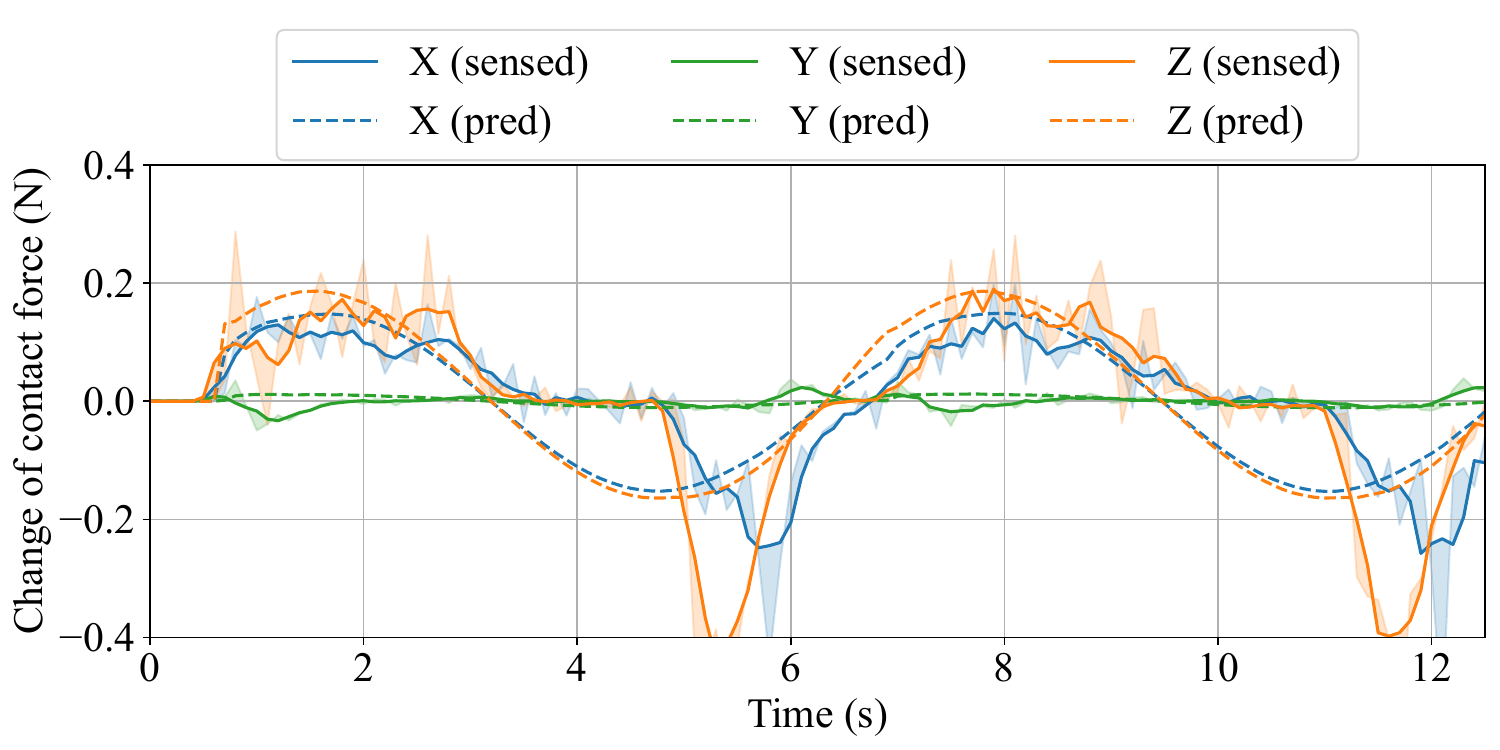} 
  \caption{\textbf{Real-world test of the motion-contact model.} The X-axis represents the normal force, while the Y- and Z-axes represent the friction forces.}
  \label{fig:exp:contact_model_real}
   \vspace{-3mm}
\end{figure}

\section{Accuracy of Motion-Contact Prediction}

The analytical motion-contact model described in Section \ref{sec:modeling} is formulated based on several assumptions:
\begin{enumerate}
    \item The manipulation process is quasi-static.
    \item The object pose is assumed to be temporally constant.
    \item Contacts are modeled as point contacts with friction, using a first-order analysis that ignores second-order contact curvatures and frictional moments.
    \item No slippage occurs temporally. 
    \item Robot joint positions are controlled by low-level PD controllers, with joint torques as the control inputs.
    \item The model of robot kinematics is accurate.
\end{enumerate}
These assumptions are not fully satisfied in real-world scenarios, making the model a coarse approximation of actual contact dynamics. Nevertheless, we find that its accuracy is acceptable to enable model predictive grasping control and achieve good performance in practice.

We design a relatively ideal experiment to evaluate the accuracy of the motion–contact model in real-world conditions. In this setup, one finger is commanded to follow a sinusoidal trajectory $\bm q_d$ while its fingertip remains in contact with a fixed object. During the motion, we record the sensed forces $\bm f^i$, the actual joint positions $\bm q$, and the commanded positions $\bm q_d$. We then compare the sensed and model-predicted force changes between consecutive steps (i.e., $\bm f^i_t - \bm f^i_{t-1}$), as illustrated in Fig.~\ref{fig:exp:contact_model_real}. 
The results show that the model achieves relatively accurate predictions when the contact forces increase (i.e., when $\bm q_d$ moves the finger further into the object).
In contrast, the measured forces behave abnormally when the finger moves back, which will be discussed in the next section.

\section{Hardware Problems} \label{sec:hardware_problems}

We find several problems of the LEAP Hand hardware in practice, which may affect the performance of grasping control in the real world. 

First, as shown in Fig. \ref{fig:exp:contact_model_real}, when $\bm q_d$ is commanded in the direction of breaking contact, the actual contact force does not decrease immediately but instead remains nearly constant and then drops abruptly.
A possible explanation for this phenomenon is related to the motor characteristics of the LEAP Hand. When the commanded change in $\bm q_d$ is small, the motors may not immediately adjust their output torques due to factors such as control deadband, backlash, or static friction in the transmission. As a result, the contact force remains nearly unchanged until the accumulated deviation exceeds a certain threshold, at which point the joint motion occurs and the contact force drops abruptly.
This phenomenon significantly affects the contact force tracking performance, particularly when the motors are commanded to move in the reverse direction to adjust frictional forces. 
As a result, we often observe that the total wrench on the object fails to fully converge before lifting, since the actual frictional forces deviate from the computed desired forces. We plan to investigate this issue further in future work, either by testing alternative hardware or by improving our formulation to mitigate such discrepancies.

Another related phenomenon is that when a human pushes a position-controlled finger, the finger exhibits a much higher resistance (characterized by $\bm K_p$) compared to when it actively pushes other objects.
This may not affect the grasping control much as the fingers are actively pressing the object during grasping. However, it affects the calibration of $\bm K_p$, as the calibration data should not be collected by manually pushing the finger.

\section{Additional Results}

\subsection{Tactile Sensing Noises} 

As for the Tac3D sensors we used in the real-world experiments, we experimentally find that the sensed normal force is relatively accurate, with relative errors below 10\%. The directions of the sensed tangential forces are accurate, and their magnitudes appear reasonable, although we have not quantitatively evaluated them. 

We conduct a simulation study to evaluate the effect of tactile sensing noises. We add gaussian noise to the sensed contact forces, and test our approach under the same setting as in Section \ref{sec:sim_shape_uncertainty}, except that a smaller test set of 3k grasps is used. 
The standard derivation of the added gaussian noise is defined as (noise scale $\times$ force magnitude) along each axis. The results are shown in Table \ref{tab:exp:tactile_noise}, where the reported normalized wrench is computed using the ground-truth contact forces. It can be seen that the grasping performance is not affected much when the noise scale is below 0.2, but undesired object movements are increase significantly at a noise scale of 0.5. It should be noted that while Gaussian noise is used for evaluation, the noise distribution in real-world tactile sensing is more complicated, and further investigation is required.

\begin{table}[tb]
\centering
\caption{\textbf{Effect of force sensing noises in simulation evaluation.}}
\label{tab:exp:tactile_noise}
\begin{tabular}{c|cccc} 
\toprule
Noise scale & SR $\uparrow$ & Pos. $\downarrow$ & Rot. $\downarrow$ & Wrench $\downarrow$ \\ 
\hline
0 & 93.1 & 2.5$\pm$4.6 & 2.4$\pm$4.0 & 0.17$\pm$0.16 \\
0.1 & 92.9 & 2.6$\pm$4.6 & 2.5$\pm$4.3 & 0.21$\pm$0.16 \\
0.2 & 92.7 & 2.9$\pm$4.7 & 2.9$\pm$4.4 & 0.25$\pm$0.16 \\
0.5 & 92.2 & 4.8$\pm$5.6 & 4.4$\pm$4.9 & 0.35$\pm$0.17 \\
\bottomrule
\end{tabular}
\end{table}

\subsection{Analysis of Real-World Manipulation Processes}


\begin{figure*} [tb]
  \centering 
  \subfigure[Open-loop]{ 
    \label{fig:exp:case_process:op}
    \includegraphics[width=0.33\textwidth]{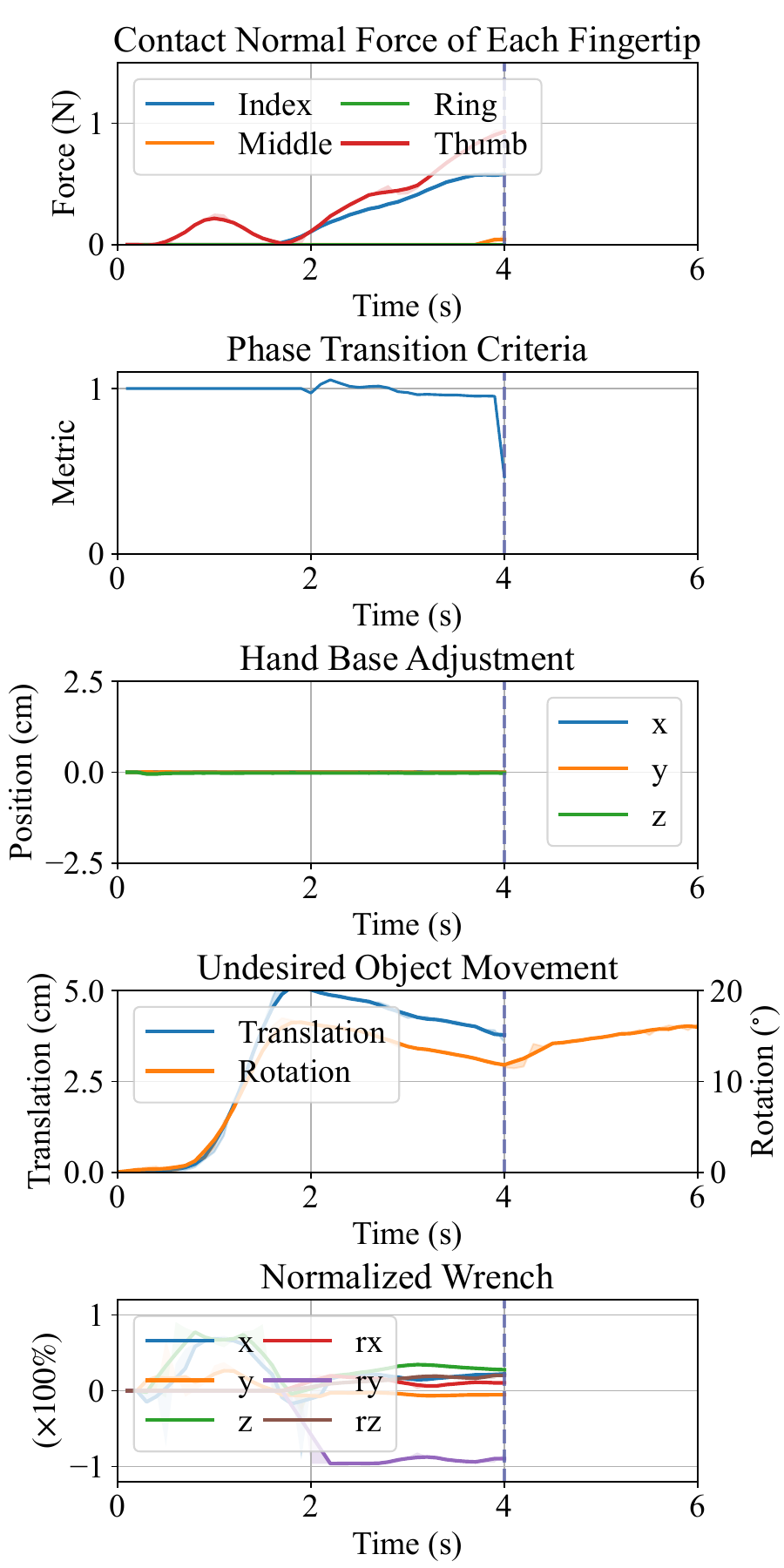} 
  } 
  \hspace{-5mm}
  \subfigure[Feedback control]{ 
    \label{fig:exp:case_process:bs1}
    \includegraphics[width=0.33\textwidth]{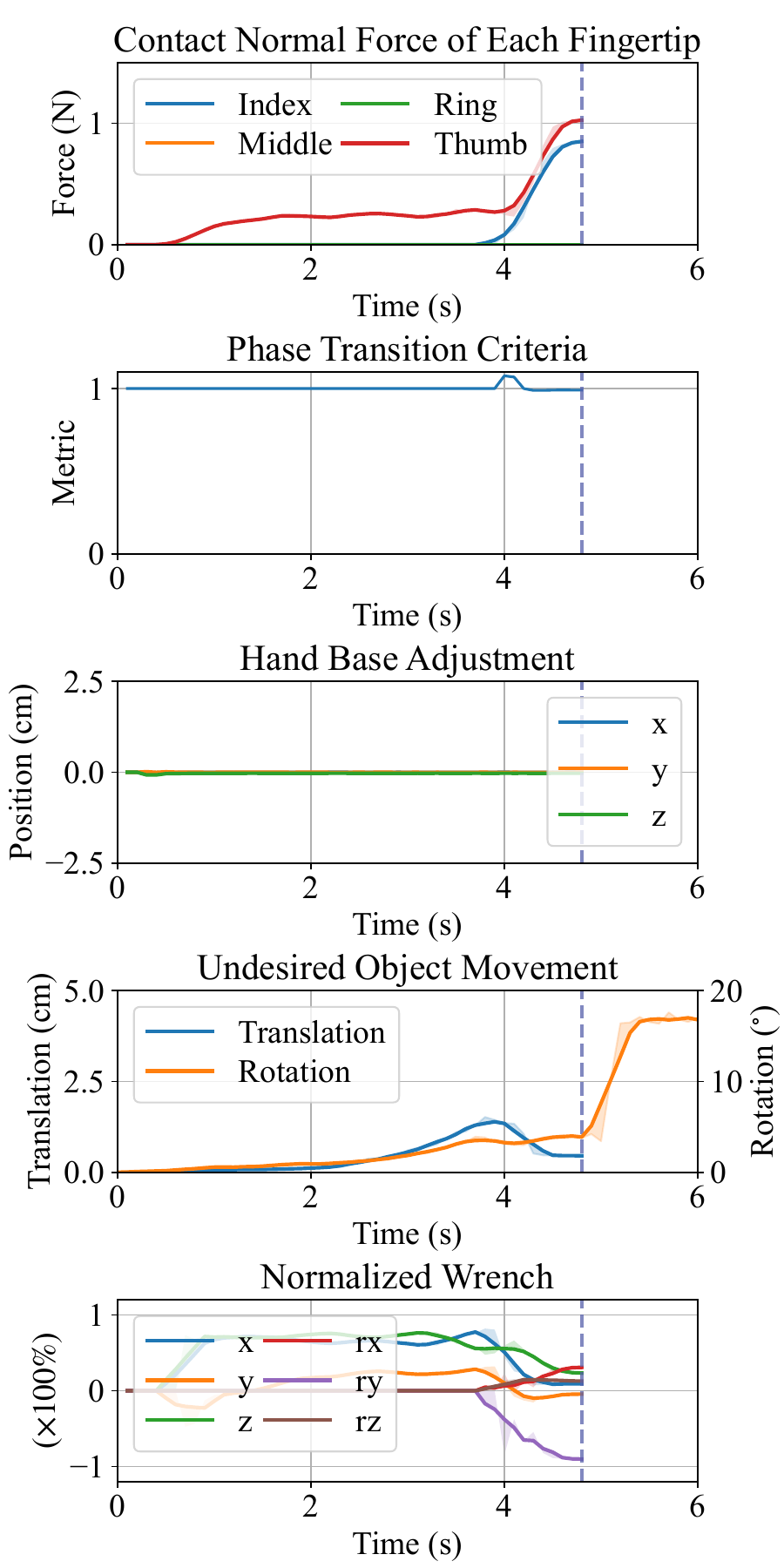} 
  }
  \hspace{-5mm}
  \subfigure[Ours]{ 
    \label{fig:exp:case_process:ours}
    \includegraphics[width=0.33\textwidth]{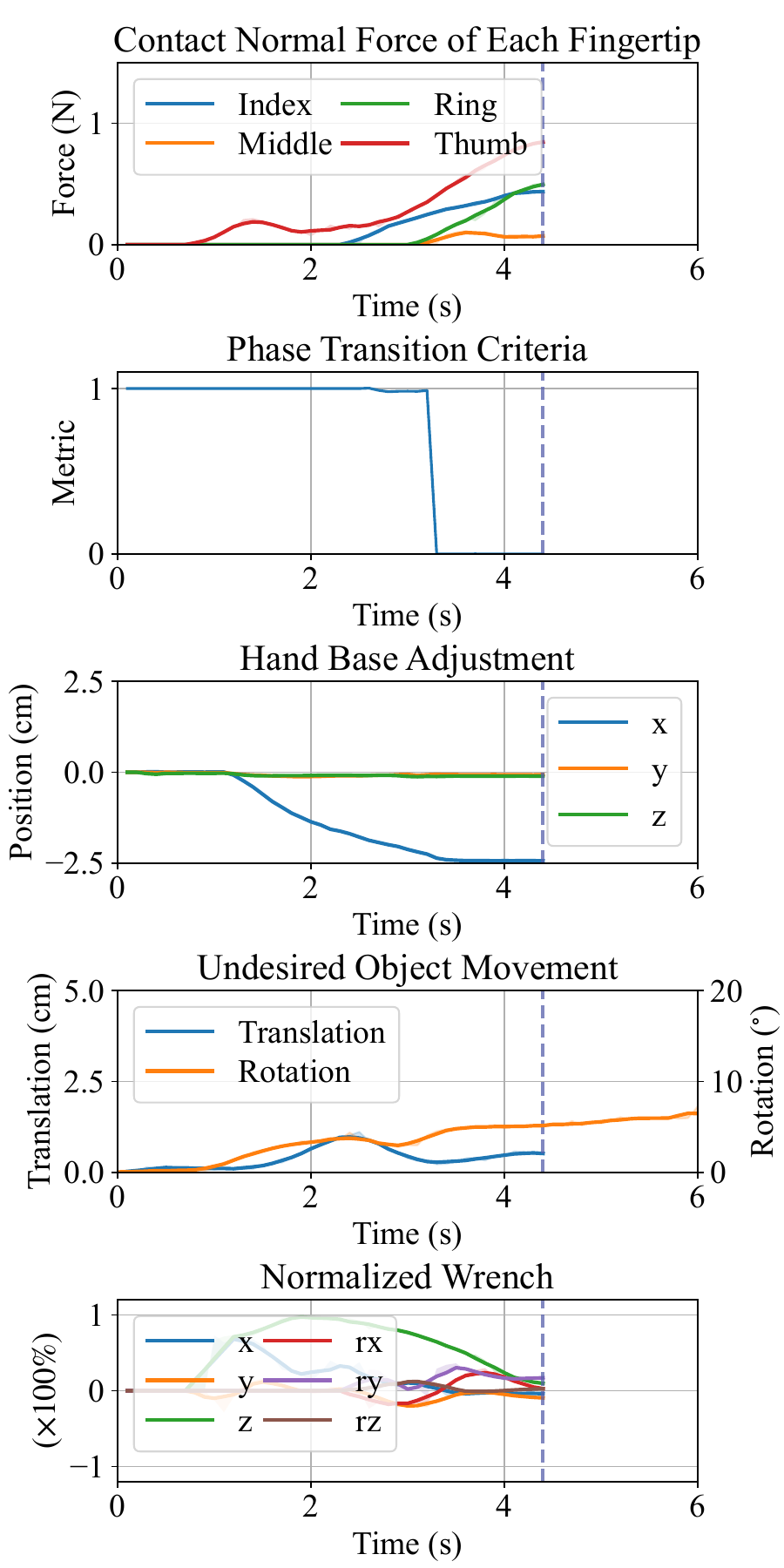} 
  }
  \caption{\textbf{Manipulation processes for grasping the position-perturbed mosquito repellent bottle} (Fig. \ref{fig:exp:real_pos_err}), comparing (a) open-loop strategy, (b) feedback control, and (c) our method. The time-series plots report contact normal forces, phase transition criteria, hand base adjustments, undesired object movements, and normalized wrenches. The vertical dotted lines indicate the lifting moment.}
  \label{fig:exp:case_process}
\end{figure*}

We provide a detailed analysis of the manipulation process for our method and the two baselines in the case shown in Fig. \ref{fig:exp:real_pos_err}. In this scenario, the mosquito repellent bottle is displaced by approximately 2 cm from its original position toward the thumb, which allows us to evaluate the performance of the methods under significant object position errors. Each method is executed five times. We visualize the recorded variables throughout the manipulation process in Fig. \ref{fig:exp:case_process}, including contact forces from each fingertip, phase transition criteria, arm adjustment motions, undesired object movements, and normalized wrenches.

Since the object is positioned close to the thumb, the thumb makes the first contact with the bottle in all methods. Afterward, our method activates arm motions to adjust the hand palm’s position to better adapt to the actual object position, whereas the baselines keep the palm fixed. During this initial phase (approximately 0 $\sim$ 2 s), the open-loop strategy moves the thumb along the planned path kinematically, which causes the bottle to be pushed and tilted. The feedback control method regulates the thumb to achieve a desired contact force; since this force is not large enough to move the bottle, the object remains stationary, but the thumb deviates significantly from the planned configuration. In contrast, our method simultaneously moves the thumb along the planned path and regulates the applied contact force on the bottle, by leveraging arm motion adjustments.

In our method, since the palm position is adjusted according to the actual object position, the index, middle, and ring fingers successfully make contact with the object while following their planned paths (approximately 2 $\sim$ 4 s), resulting in states that satisfy the phase transition criteria. In contrast, in the open-loop execution, the bottle is tilted during the initial process, preventing the middle and ring fingers from making desired contact. The feedback control method encounters the same issue: the middle and ring fingers fail to contact the bottle even at the planned final squeeze pose, because the relative position between the object and the palm deviates from the planned configuration.
The undesired grasps formed by only the thumb and index finger cannot apply balanced forces to the object. As a result, during lifting (indicated by the vertical dotted line in Fig. \ref{fig:exp:case_process}), the object rotates significantly when using the open-loop strategy or feedback control method, whereas our method maintains the object close to its initial pose.

Furthermore, in our method, the contact forces of all fingers are increased in a balanced manner during the grasping stage (approximately 3 $\sim$ 4.5 s), causing the total wrench on the object to gradually converge to zero. In contrast, the baselines generate unbalanced wrenches, which accounts for the large undesired object movements after lifting.


These manipulation processes are also visualized in the supplementary video.
The project website is at {\url{https://ada-grasp-ctrl.github.io/}}.

} 

\end{document}